\documentclass{article}
\usepackage{ijcai21}

\usepackage{times}

\usepackage{soul}
\usepackage{url}
\usepackage[hidelinks]{hyperref}
\usepackage[utf8]{inputenc}
\usepackage[small]{caption}
\usepackage{graphicx}
\usepackage{amsmath}
\usepackage{amssymb}
\usepackage{booktabs}
\usepackage{comment}
\usepackage{subfigure}
\usepackage[normalem]{ulem}
\usepackage{multirow}
\usepackage{hyperref}

\usepackage{algorithm}
\usepackage{algorithmic}
\title{Non-contact Pain Recognition from Video Sequences with Remote Physiological Measurements Prediction}
\author{
Ruijing Yang$^1$\and
Ziyu Guan$^1$\footnote{Corresponding Author}\and
Zitong Yu$^{2}$\and
Xiaoyi Feng $^3$\and
Jinye Peng $^1$\footnotemark[1]\and
Guoying Zhao$^{1,2}$\\
\affiliations
$^1$Northwest University\\
$^2$University of Oulu\\
$^3$Northwestern Polytechnical University\\
\emails
yangruijing@stumail.nwu.edu.cn, 
\{zitong.yu, guoying.zhao\}@oulu.fi, 
\{ziyuguan, pjy\}@nwu.edu.cn, 
fengxiao@nwpu.edu.cn
}

\begin{document}

\maketitle

\begin{abstract}
Automatic pain recognition is paramount for medical diagnosis and treatment. The existing works fall into three categories: assessing facial appearance changes, exploiting physiological cues, or fusing them in a multi-modal manner. However, (1) appearance changes are easily affected by subjective factors which impedes objective pain recognition. Besides, the appearance-based approaches ignore long-range spatial-temporal dependencies that are important for modeling expressions over time; (2) the physiological cues are obtained by attaching sensors on human body, which is inconvenient and uncomfortable. In this paper, we present a novel multi-task learning framework which encodes both appearance changes and physiological cues in a non-contact manner for pain recognition. The framework is able to capture both local and long-range dependencies via the proposed attention mechanism for the learned appearance representations, which are further enriched by temporally attended physiological cues (remote photoplethysmography, rPPG) that are recovered from videos in the auxiliary task. This framework is dubbed rPPG-enriched Spatio-Temporal Attention Network (rSTAN) and allows us to establish the state-of-the-art performance of non-contact pain recognition on publicly available pain databases. It demonstrates that rPPG predictions can be used as an auxiliary task to facilitate non-contact automatic pain recognition.
\end{abstract}

\section{Introduction}
Pain recognition is paramount for medical diagnosis and treatment. The ``golden standard'' of measuring pain intensity in clinics is self-report, where it is often accomplished by tools such as visual analog scales (VAS). However, such measurement is not applicable in the situation where it is not possible to communicate, such as newborns, patients in unconscious states, or patients with verbal or mental impairment. Besides, self-report suffers from subjective biases. Therefore, automatic recognition of pain is crucial for alleviating clinicians' workload, providing references for actual treatments, and raising objectiveness in recognition. Facial expressions and physiological changes (e.g., heart, brain or muscular activities) are two sources of information that are correlated with pain. 

Existing methods for automatic pain recognition fall into three categories: (1) quantifying facial appearance changes by videos; (2) exploiting correlated physiological cues; (3) the combination of the above two measurements in a multi-modal manner. For the first category, many works focused on developing effective facial features to quantify pain, where both spatial information and temporal dynamics are considered locally \cite{werner2016automatic,tavakolian2020self}. However, appearance changes are easily affected by personality or past experiences, which would hinder objective pain recognition. Moreover, those methods fail to capture long-range dependencies, e.g., the complete evolution of an expression in the temporal dimension and the correlation between eyes and mouth regions in the spatial dimension. Compared to appearance changes, physiological signals can provide relatively more objective cues. This motivates works of the second category  \cite{kachele2015multimodal,lopez2017multi,lopez2018continuous,thiam2019exploring}, which focused on determining the relationship between pain and different physiological modalities by utilizing predefined statistical or frequency traits. Nevertheless, the physiological signals are not readily available: they require professional medical apparatus correctly configured and with sensors attached to the subject's body under strict instructions. It would be hard to meet these requirements in common scenarios (e.g., home caring), and the sensors increase the discomfort of subjects. There were efforts to take both facial appearance changes and physiological cues into consideration for better performance~(the third category, e.g., \cite{werner2014automatic,kachele2015multimodal}), but the inconvenience drawbacks discussed above still apply.

Non-contact, vision-based physiological measurements prediction attracts much attention in the computer vision field as it provides convenient assessment and potentials for unobtrusive measurements in ambient intelligence \cite{DeepPhys}. Remote photoplethysmography (rPPG) is one of such methods which aim to recover heart activities from video sequences \cite{sun2015PPG}. The principle is based on the subtle changes in light reflected from the skin in certain ROIs which are caused by the changes in blood volume. Such heart activities are proven to be correlated with pain \cite{terkelsen2005acute}. Recovery of vital physiological signals from painful faces would broaden the potential applicable scenarios of automatic pain recognition from clinical use to more common cases such as long-term monitoring of the seniors at home. However, no previous work has explored leveraging this idea for automatic pain recognition.

A straightforward idea of combining rPPG and appearance changes is to treat rPPG signals as real physiological measurements and follow \cite{werner2014automatic} to concatenate high-level features from the two modalities. However, this scheme loses temporal and spatial structural information which would be important for pain recognition. In this paper, we propose a non-contact rPPG-enriched Spatio-temporal Network for accurate pain recognition, where only video data is utilized to extract facial dynamics and recover the rPPG signals. We incorporate physiological cues differently - via the proposed Visual Feature Enrichment module (VFE) to enrich the facial representation such that rPPG-related ROIs are considered to complement facial representation learning with emphases on certain time steps. Besides, we also improve appearance-based learning by designing a Spatio-Temporal Attention module (STA) on top of 3D ConvNet where both local dynamics and long-range dependencies in space and time can be captured.

Our contribution can be summarized as follows: (1) we propose a non-contact, end-to-end architecture for pain recognition named rPPG-enriched Spatio-temporal Attention Network (rSTAN). rSTAN builds an effective video-level facial representation by exploiting rPPG signals recovered from video data to enrich the facial representation, without introducing extra input sources; (2) we design a Visual Feature Enrichment module (VFE) to distill the physiological information to complement and guide the visual representation learning; (3) we propose a Spatio-Temporal Attention module (STA) to capture long-range dependencies in both space and time during the evolution of pain expression; (4) Extensive experiments on the benchmark databases demonstrate that our approach achieves the state-of-the-art performance, and indicate that physiological signals prediction can be viewed as an auxiliary task for automatic pain recognition.

\section{Related Work}
Many advances have been made on developing effective visual features to quantify pain based on appearance changes. In \cite{werner2014automatic,werner2016automatic}, the statistical features of landmark distance changes and facial texture changes were analyzed and used as facial behavioral descriptors to classify different pain intensities. On the same task, the temporally extended version of Local Binary Pattern (LBP-TOP) was employed to capture the facial dynamics \cite{yang2016pain}. In \cite{zhi2019multi}, geometric and texture changes, as well as head movements were encoded as facial representations for pain estimation. 

Physiological signals are also considered to be helpful for pain recognition. Specifically, changes in heart rate, skin conductance, brain hemodynamics, skeletal muscles, that can be measured by electrical devices, were exploited. 
In \cite{lopez2017multi}, a multi-task network was utilized to estimate self-reported pain with pre-defined physiological features as input. In their later work \cite{lopez2018continuous}, they further exploited the relationship between physiological cues and pain. With similarly pre-defined physiological features, a LSTM-NN was used to classify different pain intensities against no pain. Only recently, an end-to-end approach replaced those handi-crafted features of ECG, EDA, EMG signals \cite{thiam2019exploring}. However, the informative visual cues are not exploited in these studies.

\begin{figure*}[ht]
\centering
\includegraphics[scale=0.4]{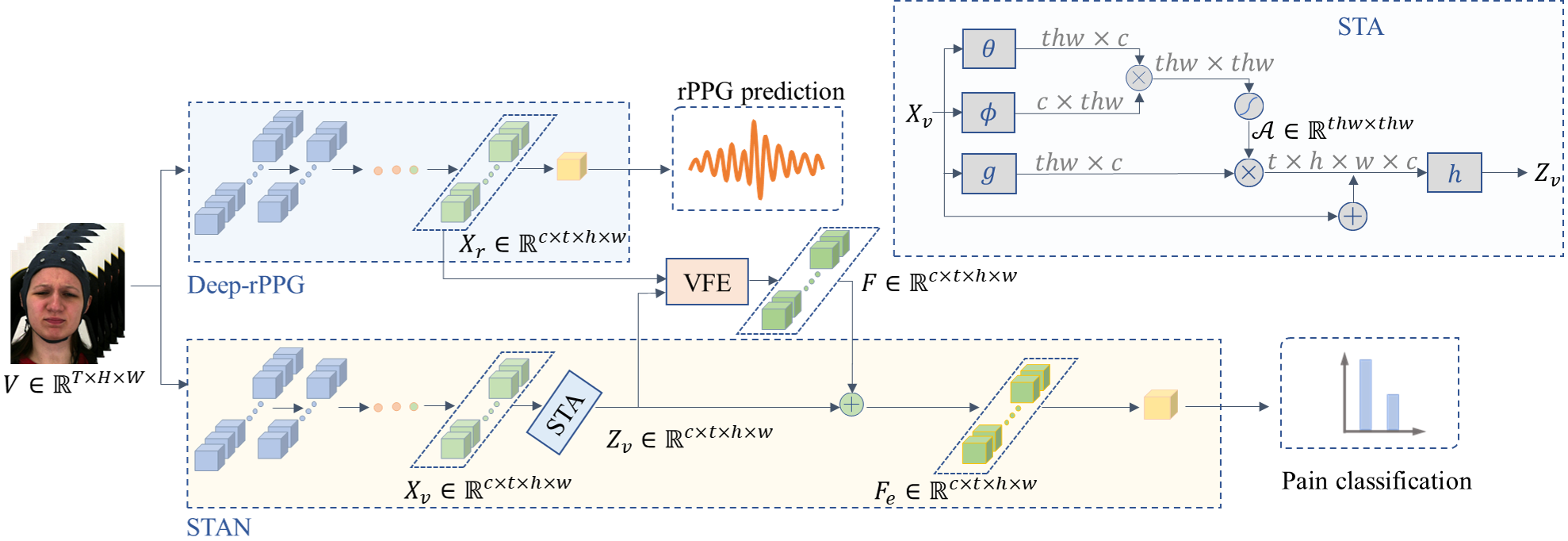}
\caption{The proposed rPPG-enriched Spatio-temporal Attention Network (rSTAN) and  the Spatio-temporal Attention module (STA). The details of STA is illustrated on the upper right of the figure, where \(\oplus\) represent element-wise addition, \(\otimes\) refers to matrix multiplication.}
\label{framework}
\end{figure*}

Deploying physiological data to assist facial appearance-based pain recognition is becoming the trend in recent years. In \cite{werner2014automatic}, three types of physiological signals (electrocardiogram, electromyography, galvanic skin response) were combined with facial features to classify the severity of pain in a multi-modal way. In \cite{kachele2015multimodal}, facial appearance features and physiological features were fused for pain estimation via different fusion schemes. 

Remote photoplethysmography (rPPG) prediction is a rapidly growing research topic, for its potential application in heart rate estimation \cite{DeepPhys}, respiration detection \cite{van2016robust} and face anti-spoofing \cite{liu2018remote}. Current studies on this topic focus on recovering rPPG signal and deploying it straight to downstream tasks. However, such direct employment provides limited information for pain recognition. In this work, we treat rPPG prediction as an auxiliary task to facilitate the facial representation learning.

It can be seen from previous studies that involving more sources of information improves the performance for pain recognition. Facial appearance-based methods suffer from the subjectiveness issue. Though physiological measurements can alleviate this issue, it requires considerable device supports and human interference and causes discomfort to the subjects. Hence, combining physiological cues with facial representations in a non-contact manner could significantly advance the technical level of automatic pain recognition. 
\section{rPPG-enriched Spatio-Temporal Attention Network (rSTAN)}
Given a video snippet \(\mathbf{V}\in\) $\mathbb{R}^{T\times H\times W}$, our rSTAN processes it into two branches: the rPPG recovery branch (Deep-rPPG) and the facial representation learning branch (STAN), where \(T, H, W\) represent the number of frames, image height and width respectively. The objective of our framework is to learn an enriched facial representation $\mathbf{F}_e\in \mathbb{R}^{c\times t\times h\times w}$ that can distinguish different pain intensities, where $c$ denotes the number of channels, $t, h, w$ are the temporal length, height and width of the feature maps. As shown in Figure \ref{framework}, the two branches exploit similar architectures where 3D ConvNet is used as backbone. Hence, they could capture both facial and rPPG dynamics during different pain states. To capture distant correlations spatially and temporally, in the facial representation branch, we propose a Spatio-Temporal Attention module (STA) which is applied on facial feature maps $\mathbf{X}_v\in\mathbb{R}^{c\times t\times h\times w}$ to obtain the attended feature maps $\mathbf{Z}_v$ of the same size. To facilitate effective facial feature learning, we design a Visual Feature Enrichment module (VFE) to incorporate rich information gained from the rPPG branch, where both rPPG-related spatial ROIs and dynamics are exploited as guidance. Taking rPPG feature maps $\mathbf{X}_r\in\mathbb{R}^{c\times t\times h\times w}$ as input, the objective of VFE is to obtain a facial representation $\mathbf{F}\in\mathbb{R}^{c\times t\times h\times w}$ by computing a same-shape attention map $\mathbf{M}$ obtained from $\mathbf{X_r}$ to enrich $\mathbf{X}_v$ in a way that pain-related dynamics can be sufficiently focused on. Finally, to obtain the enriched facial representation $\mathbf{F}_e$, a residual connection is applied to allow the gradient flow. Formally, the enriched facial representation is expressed as:
\begin{align}
    \mathbf{F}_e = \mathbf{F} + \mathbf{Z}_v,
\end{align}
where the operators $``+''$ represents residual connection. 

\subsection{Spatio-temporal Attention}
\label{sec:sta}
Facial dynamics play an important role in assessing pain. A spontaneous facial expression often lasts between 0.6 second to 4 seconds \cite{ekman2003darwin}. Thus, dynamics captured by local operators are not sufficient to describe the evolution of an expression. Besides, 3D convolution usually treats all frames equally and is computed locally, thus cannot capture more global information with focus. Therefore, in this paper, we propose a spatio-temporal attention network (STAN), which uses 3D convolutions as the basis with a Spatio-Temporal Attention module (STA) to capture correlations in distant facial regions and frames attentively. Such an attention module is used to compute the response at a position in the input feature maps by considering all the other positions. Thus the correlations among distant positions are measured. 

Here, we re-denote the facial feature maps as $\mathbf{X}_v=(\mathbf{x}^1,\mathbf{x}^2,...\mathbf{x}^i,...,\mathbf{x}^n)$, where $\mathbf{x}^i\in\mathbb{R}^c$, $i\in\{1,2,...,n\}$ and $n=t\times h \times w$.
The attention map $\mathcal{A}\in\mathbb{R}^{n\times n}$, in STAN can be described as:
\begin{align}
    \mathcal{A}(\mathbf{X}_v) = \mathcal{S}({\theta{(\mathbf{X}_v)^T}\phi{(\mathbf{X}_v)}}),
\label{eq:A}
\end{align}
and the attended facial feature maps \(\mathbf{Z}_{v}\) can be obtained via:
\begin{align}\label{att_op}
    \mathbf{Z}_{v} = h(\mathcal{A}(\mathbf{X}_v)g(\mathbf{X}_v)+\mathbf{X}_v), 
\end{align}
where \(\mathcal{S}\) in Eq.(\ref{eq:A}) represents the \textit{softmax} function. The function \(g\) computes a representation of the input feature and function \(h\) computes a new embedding after attention. \(\theta(\mathbf{X}_v) = W_{\theta}\mathbf{X}_v\) and \(\phi(\mathbf{X}_v) = W_{\phi}\mathbf{X}_v\) are the two embeddings to compute the similarity in an embedding space. With $\theta$ and $\phi$, the attention module is able to learn the mutual relationship between any two points in the whole spatial-temporal space via matrix multiplication. Thus the correlation among distant regions such as eyes and mouth can be measured. This also applies to any regions in temporally distant frames. Such behavior is due to the fact that all the positions $\forall{i}$ in the feature maps are considered. The detailed structure of STA is illustrated in the upper right part of Figure~\ref{framework}.

The attention module is inserted into the high-level feature maps in our network with dimensions of \(t = 16, h = w = 7\). This is due to the high computational cost at lower level feature maps, whereas at the high level it is comparable to typical convolution operations in standard neural networks. 

\begin{figure*}[ht]
\centering
\includegraphics[scale=0.417]{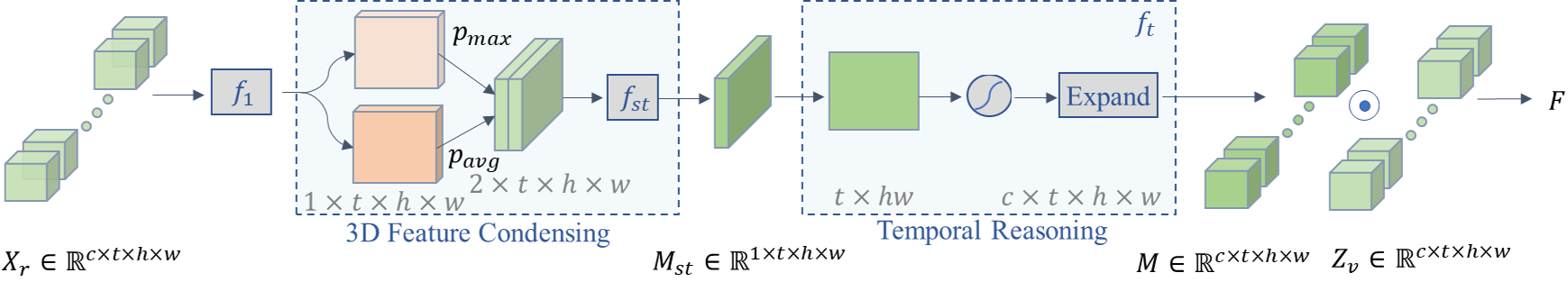}
\caption{The proposed Visual Feature Enrichment module (VFE). The sizes of feature maps at different stages are stated in gray font and variables. \(\odot\) represent element-wise multiplication. \(f_1, f_{st}\) are implemented as \(1\times 1\times 1\) convolution and \(f_t\) refers to softmax function.}
\label{fig:vfe}
\end{figure*}

\subsection{Visual Feature Enrichment}
According to the principle of rPPG recovery, capturing subtle light reflection changes from facial skin in ROIs naturally enables rPPG features to be spatial attention on regions that may complement the facial representation in STAN. Such ROIs usually are the regions that have strong pulsatile flows such as cheeks and foreheads, where the visual changes are subtle and difficult to detect via normal facial recognition systems, but vital indicators for pain recognition (e.g. cheek areas, corresponding to Action Unit 6 in Facial Action Coding System). 
Along the temporal dimension, heart rate changes are also correlated with pain which may cause facial expression changes. The temporal dynamics of the rPPG features that describe the heart rate evolving over time would implicitly indicate where in time the pain expression may occur. Therefore, rPPG features could guide spatially and temporally the learning of facial representations to be more focused on important cues that may be overlooked by pure appearance-based methods. To this end, we propose a Visual Feature Enrichment module (VFE). VFE attempts to complement facial representation by considering the spatial ROIs from rPPG features with focuses on important time steps that contain physiological variations related to pain.

As illustrated in Figure~\ref{framework} and Figure~\ref{fig:vfe}, we take a 4D feature map \(\mathbf{X}_r \in\) $\mathbb{R}^{c\times t \times h \times w}$ from the latter layer of rPPG branch as the input of VFE, since it contains high-level semantics that related to rPPG signals. A new feature embedding is computed via function $f_1$, which is implemented as a $1\times 1\times 1$ convolution. Then, different from simple max pooling, we apply both average pooling and max pooling along the channel axis to obtain the important and average component of the rPPG signals, i.e. two compact 3D feature maps, denoted as $\mathbf{p}_{avg}$ and $\mathbf{p}_{max}\in \mathbb{R}^{1\times t\times h\times w}$ (indicated by the pink boxes in Figure \ref{fig:vfe}). The reason is that for physiological measurements, combining an important signal and a less important one can produce higher signal to noise ratio than only using the most important one. We then concatenate the two maps and deploy a new function $f_{st}$, also implemented as a $1\times 1\times 1$ convolution, to output a condensed spatio-temporal feature map $\mathbf{M}_{st}$ that describes the dynamics of rPPG-related ROIs. To incorporate the temporal information that describes how rPPG-related ROIs evolves over time, we further perform temporal reasoning on $\mathbf{M}_{st}$ with a softmax function $f_{t}$ applied along the temporal dimension to obtain the temporal attention map (right part of Figure~\ref{fig:vfe}). Then it is expanded along the channel axis as $\mathbf{M}$. In such a way, the temporal importance is highlighted for each spatial position on the feature map. Thus $\mathbf{M}$ can be used to guide facial representation learning with attention distributed throughout time. We summarize the 3D feature condensing and the temporal reasoning as:
\begin{equation}
    \mathbf{M}_{st} = f_{st}([\mathbf{p}_{avg}, \mathbf{p}_{max}]) \in\mathbb{R}^{1\times t\times h\times w}
\end{equation}
\begin{equation}
    \mathbf{M} = e(f_t(\mathbf{M}_{st})) \in\mathbb{R}^{c\times t\times h\times w},
\end{equation}
where \([\cdot, \cdot]\) stands for feature concatenation along the channel axis, and $e$ refers to the channel expansion. 

Finally, the enriched feature map $\mathbf{F}$ is yielded by applying the attention map $\mathbf{M}$ on the attended facial representation $\mathbf{Z}_v$:
\begin{equation}
    \mathbf{F} = \mathbf{Z}_v \odot \mathbf{M}, 
\end{equation}
where the operator \(\odot\) refers to the Hadamard product.

The proposed VFE prefers to modulate the spatio-temporal visual features with emphasis guided by the physiological cues, which is different from the traditional feature fusion paradigm (e.g., feature concatenation and summation).

\subsection{Deep-rPPG for Pulse Signal Recovery}
\label{sec:rPPG}
We propose Deep-rPPG, a 3D ConvNet-based temporal encoder-decoder network for rPPG signals recovery from facial videos. Inspired by the Inflated 3D ConvNet (I3D) \cite{carreira2017quo} which exploits strong spatio-temporal cues for video understanding, the proposed Deep-rPPG adopts similar backbone but with the following changes: 1) the temporal strides of the last Max-Pooling and Avg-Pooling layers are set to one to preserve more temporal information; and 2) two deconvolution operations for temporal feature decoding are utilized for the rPPG recovery. As a result, Deep-rPPG is able to capture rich spatio-temporal context features and recover stable rPPG signals. 
Negative Pearson correlation is adopted as the loss function to maximize the similarity in signal tendency.
\begin{equation}
\resizebox{.91\linewidth}{!}{$
    \displaystyle
    \mathcal{L}_{rPPG}=1-\frac{T\sum_{1}^{T}xy-\sum_{1}^{T}x\sum_{1}^{T}y}{\sqrt{(T\sum_{1}^{T}x^2-(\sum_{1}^{T}x)^2)(T\sum_{1}^{T}y^2-(\sum_{1}^{T}y)^2)}},
$}
\end{equation}
where \(T\) is the length of the signals, \(x\) is the predicted rPPG signals, and \(y\) is the ground truth: smoothed ECG signals.

\section{Experiments and Results}
\subsection{Databases and Evaluation Metrics}
\paragraph{Databases.} We evaluate the performance of the proposed method on: the BioVid Heat Pain Database (the \textbf{BioVid} for short) \cite{walter2013biovid} and the UNBC-McMaster Shoulder Pain Expression Archive Database (the \textbf{UNBC} for short) \cite{LuceyUNBCdatabase}. \textbf{The BioVid} is a multi-modal pain database with both video and physiological data. A total number of 8700 samples from 87 subjects were collected and equally distributed in 5 classes: T0-T4, where T0 refers to no pain and T4 is the tolerance threshold. Pain stimuli were introduced to subjects on their right arm using thermodes. \textbf{The UNBC} is collected from volunteers who were self-identified as patients suffering from shoulder pain. In total, 200 video sequences containing spontaneous facial expressions were recorded from 25 subjects. Videos were recorded when the subjects were experiencing a series of active and passive motion of their affected and unaffected limbs. In this paper, the UNBC is used for cross-database validation, since the physiological data are unavailable in this database.

\paragraph{Performance Metrics.} Results are reported in terms of accuracy for distinguishing the tolerance threshold of pain (T4) from no pain (T0) and five discrete pain intensities (T0-T4). For the cross-database test, Area Under Curve (AUC) is used as the evaluation metric. 

\subsection{Implementation Details}\label{sec:Implementation}
For all experiments conducted on BioVid, we omit the first 10 frames of each video according to the visual inspection that subjects are prone to have expression changes during the latter part of the video and so do the ECG signals. Our framework is built upon the Inflated 3D ConvNet \cite{carreira2017quo} and pretrained using CASIA-WebFace dataset \cite{yi2014casia}. For the parameter selection, we randomly split all the 87 subjects into 5 folds and use 5-fold cross-validation to determine the best parameters. While for the comparisons to the state of the arts, we follow the leave-one-subject-out protocol (LOO) as previous works \cite{werner2014automatic,werner2016automatic}. MTCNN \cite{zhang2016mtcnn} is utilized to detect and crop the facial area.
The proposed method is trained on Nvidia P100 using PyTorch. For the input videos, the original video is downsampled to $L=64$ since it produces the best performance. Adam is used as the optimizer with a learning rate of 2$e-$4 which is decayed after 10 epochs with a multiplicative factor $gamma$=0.8. To learn the parameters of the two branches more efficiently, we train them separately and fine-tune the whole framework jointly. 
\subsection{Comparison to Uni-physiological Modality}\label{sec:rppgExp}
A reliable rPPG representation is crucial for the VFE in rSTAN. On one hand, the rPPG signals recovered from Deep-rPPG should be highly correlated with the ground-truth smoothed ECG signal. On the other hand, rPPG-based pain recognition should achieve comparable performance to ECG-based approaches. To this end, given the recovered rPPG signals, a 1D-CNN is utilized for pain-derived physiological feature extraction and pain recognition in an end-to-end fashion. For rPPG recovery, an illustrative comparison between the recovered rPPG and smoothed ECG signal in both pain and no pain states is presented in \textsl{Appendix}\footnote{Readers may refer to: \textit{\url{https://arxiv.org/abs/2105.08822}} for the Appendix}. For pain recognition, we compare the results to the literatures which use the sensor-collected ECG signals as uni-modality to extract predefined features. As shown in Table \ref{tab: ecgcmp}, the proposed Deep-rPPG achieves comparable performance to the sensor-based methods. This is as expected since we only use the video data as the input. In comparison, our method outperforms frequency features used in \cite{lopez2018continuous} which states the effectiveness of the rPPG recovery process. On behalf of deep-learning based method, the proposed spatio-temporal Deep-rPPG also competes the one using 2D-CNN \cite{thiam2019exploring}.

\begin{table}[hb]\addtolength{\tabcolsep}{-5pt}
\centering
\resizebox{\columnwidth}{!}{%
\begin{tabular}{lrr}
\toprule
 Method & Acc (\%) & Validation\\
 \midrule
 ECG-Random Forest \cite{werner2014automatic}     & \underline{62.00} & LOO \\
 ST-NN \cite{lopez2017multi} & 59.32 & 10 folds \\
 LF \cite{lopez2018continuous}  & 57.69 & LOO \\
 HF \cite{lopez2018continuous} & 50.10 & LOO \\
 LF/HF \cite{lopez2018continuous} & 53.15 & LOO \\
 ECG-CNNs \cite{thiam2019exploring} & 57.16 & LOO \\
 Deep-rPPG (ours) & \textbf{58.92} & LOO\\
 \bottomrule
\end{tabular}
}
\caption{Comparison to uni-physiological modality}
\label{tab: ecgcmp}
\end{table}
\subsection{Comparison to Appearance-based Approaches}
For the appearance-based approaches, we compare STAN and rSTAN to previous studies whose facial representations are derived purely from video data, to demonstrate the effectiveness of our STA and VFE modules. This is because the proposed rSTAN only uses video data as input, as same as those appearance-based approaches. The STAN without the attention module is also compared. From Table~\ref{tab:appearance}, it is clear that the proposed STAN surpasses its version without the attention module by a margin of 2.3\% for binary and 1.8\% for five class classification, indicating that the proposed attention module is able to provide complementary dynamic information to 3D ConvNet via the correlation computation in STA. When incorporating Deep-rPP branch via VFE, the performances are further boosted by 0.6\% and 0.4\% for both tasks. This reveals that the recovered rPPG signals can further complement the appearance representations and can be used as an auxiliary source to facilitate pain recognition. 

The comparison to all previous studies shows that our STAN can capture global information by taking the distant relationship among facial regions and frames into consideration, hence outperform those appearance-based methods by a large margin. In addition, the comparison also demonstrates that the 3D ConvNet is more effective in capturing facial dynamics than those statistical or geometric based facial dynamics. It is worth noting that the facial features with head movements \cite{werner2014automatic,werner2016automatic} are not only using video data, but also exploiting the depth map collected from a Kinect camera to generate head movement features.

\begin{table}[hb]\addtolength{\tabcolsep}{-5pt}
\centering
\resizebox{\columnwidth}{!}{%
\begin{tabular}{lrr}
\toprule
Method  & Acc (bin) & Acc (all) \\
\midrule
Facial Expression \cite{werner2014automatic}    & 70.8   &  -  \\
FE + Head movement \cite{werner2014automatic}      & 71.6   &  -  \\
Texture features \cite{werner2016automatic}      & 68.7   &  28.9  \\
3D dist \cite{werner2016automatic}              & 72.1    &  30.3 \\
FE + Head movement \cite{werner2016automatic}      & 72.4 & 30.8 \\
GME \cite{zhi2019multi}    & 68.0  & 25.0 \\
Distance \cite{zhi2019multi} & 70.0 & 29.0 \\
SSD \cite{tavakolian2020self} & 71.0 & - \\
STAN-without-attention (ours)                                & 76.0 & 32.8 \\
STAN (ours)                                  & 78.3 & 34.6 \\
rSTAN (ours)                                 & \textbf{78.9} & \textbf{35.0}\\
\bottomrule
\end{tabular}%
}
\caption{Comparison to appearance-based methods. ``Acc(bin)" and ``Acc(all)" refer to binary and five intensity classification.}
\label{tab:appearance}
\end{table}

\subsection{Comparison to Multi-modal Approaches}
The multi-modal approaches here refer to those using both video and physiological signals as input and their fusion for pain recognition. The proposed rSTAN is not a purely multi-modal approach, since only video data are used as the network input. There is no similar multi-task framework in literature, thus we compare the results to those with multi-modal fusion schemes. All the compared methods are using early fusion. This is because rSTAN also performs at feature level, not decision level. In Table \ref{tab:multi}, it is interesting that rSTAN outperforms the method \cite{werner2014automatic}, even though they took videos, depth maps, and physiological signals as inputs, whereas ours only utilize video data. This is mainly due to the superior performance of STAN compared to their facial representation (FE $+$ Head movement) in Table~\ref{tab:appearance}.

rSTAN does not outperform \cite{kachele2015multimodal} and \cite{zhi2019multi} for the following two reasons: 1) we only use video as input, whereas all those methods utilize video and three types of physiological signals which contain richer pain related cues; 2) our physiological signal is recovered from video data, but theirs are collected from attached sensors, hence there will be a gap between the recovered signal and sensor collected signal. This explains the small improvement from STAN to rSTAN, compared to previous approaches' significant improvements after involving physiological signals. Meanwhile, it also indicates that if we incorporate more physiological cues, the performance would be further enhanced. Since our focus is to realize non-contact pain monitoring, contact-based physiological signals are not considered at the current phase. 

Furthermore, we also compare the proposed VFE to the traditional early fusion methods. We implemented two versions of early fusion for comparison, as listed in the lower part of Table~\ref{tab:multi}, where we: 1) concatenate the flattened features from the two branches (``Early Fusion Flatten''); 2) directly concatenate feature maps $\mathbf{X}_r$ and $\mathbf{Z}_v$ from the two branches (``Early Fusion''). When the VFE is replaced by the flattened version, the performance degenerated. This is mainly due to the intra-correlated spatio-temporal information between the two branches is discarded. Compared to direct feature maps concatenation, rSTAN still outperforms it. This indicates that incorporating rPPG information in an attentive way is more effective than simple concatenation and further demonstrates that rPPG can be used as an auxiliary source for pain recognition.
\begin{table}[ht]\addtolength{\tabcolsep}{-5pt}
\centering
\resizebox{\columnwidth}{!}{%
\begin{tabular}{llrr}
\toprule
Modality & Method  & Acc(bin) & Acc(all)\\
\midrule
\multirow{3}{4em}{MS+\\Video} & FE+Head+Bio \cite{werner2014automatic}  & 77.8 & - \\
& FE+Bio \cite{kachele2015multimodal} & 82.7 & - \\
& FE+Head+Bio \cite{zhi2019multi}    & 85.0 & 39.0  \\
\midrule
\multirow{3}{4em}{Video} & Early Fusion Flatten (ours) & 78.1 & 34.2\\
                         & Early Fusion (ours) & 78.2 & 34.4\\
                         & rSTAN (ours)  & 78.9 & 35.0 \\
\bottomrule
\end{tabular}%
}
\caption{Comparison to the state-of-the-art multi-modal approaches. ``MS+Video" represents utilizing multiple physiological signals and videos as input and ``Video'' refers to using only video data as input. ``FE", ``Head", ``Bio" represent facial appearance, head movement and physiological features.}
\label{tab:multi}
\end{table}
\subsection{Cross-database Validation}
Cross-database validation is to test the generalization ability and robustness of the proposed method. However, it is rarely tested on the task of automatic pain recognition. This may be due to the degenerated performance when the well-trained model is used for testing on a different database \cite{othman2019crossdb}. In this section, we conduct cross-database validation on UNBC to test the generalization ability of the proposed method. The model learns the rPPG-enriched facial representations from the BioVid, and the same representations are extracted from UNBC for testing. It is expected that the performance drops, compared to the methods trained on UNBC. The main reason would be the subjective nature of pain, which further leads to the exaggerated gap between different database domains in the cross-database test settings than other tasks. Similar trend was also demonstrated in \cite{tavakolian2020self} where the performance of BioVid-trained SSD on UNBC (SSD-crossDB) dropped significantly. Such findings also highlighted the importance of incorporating physiological cues. Interestingly, we found that our rSTAN improves the performance on the UNBC database, compared to STAN, further stating that our VFE enriches the facial representation.

\begin{table}[ht]\small
\centering
\resizebox{\columnwidth}{!}{
\begin{tabular}{lr}
\toprule
Method  &  AUC \\
\midrule
SPTS + CAPP (with PCA) \cite{werner2016automatic}$\ast$ & 91.7 \\
Align crop+LSTM \cite{pau2017deep}$\ast$ & 91.3 \\
SSD \cite{tavakolian2020self}$\ast$ & 83.2 \\
SSD-crossDB \cite{tavakolian2020self}$\star$ & 69.2 \\
STAN (ours)$\star$ & 62.8 \\
rSTAN (ours)$\star$ & 72.3 \\
\bottomrule
\end{tabular}
}
\caption{Cross-Database Results on UNBC database. The symbol $\ast$ represents methods trained on UNBC database and symbol $\star$ refers to the cross-database test results on UNBC database.}
\label{tab:unbc}
\end{table}
\section{Conclusions}
This paper presents a non-contact rPPG-enriched Spatio-temporal Attention Network (rSTAN) to obtain an enriched facial representation for pain recognition by only taking video data as input. A Visual Feature Enrichment module (VFE) is proposed to enrich the facial representation by exploiting physiological cues (rPPG) recovered from video, where a Deep-rPPG is designed to recover rPPG signals as an auxiliary task. Besides, a Spatio-Temporal Attention module (STA) is proposed to model long-range dependencies of the facial representations. Extensive experimental results demonstrate the effectiveness of our framework and the proposed VFE, STA and Deep-rPPG. Moreover, it also shed light on the potential application in non-contact pain monitoring and more broadly ambient intelligence. 

\section*{Acknowledgments}
We thank Wei Peng, Youqi Zhang, Henglin Shi as well as Long Chen for their help, support and valuable discussions on this project. This work was partly supported by the Xi’an Key Laboratory of Intelligent Perception and Cultural Inheritance (No. 2019219614SYS011CG033), the National Natural Science Foundation of China (No. 61936006, 61772419 \& 6177051263), the Program for Changjiang Scholars and Innovative Research Team in University (No. IRT\_17R87), and Natural Science Basic Research Program of Shaanxi (No. 2020JQ-850 \& 2019JM-103). This works was also supported by the Academy of Finland for ICT 2023 project (grant 328115), Infotech Oulu, and project MiGA (grant 316765).

\clearpage
\bibliographystyle{named}
\bibliography{ijcai21}

\begin{thebibliography}{}

\bibitem[\protect\citeauthoryear{Carreira and
  Zisserman}{2017}]{carreira2017quo}
Joao Carreira and Andrew Zisserman.
\newblock Quo vadis, action recognition? a new model and the kinetics dataset.
\newblock In {\em Proc.of CVPR}, pages 6299--6308, 2017.

\bibitem[\protect\citeauthoryear{Chen and McDuff}{2018}]{DeepPhys}
Weixuan Chen and Daniel McDuff.
\newblock Deepphys: Video-based physiological measurement using convolutional
  attention networks.
\newblock In {\em Proc. of ECCV}, pages 349--365, 2018.

\bibitem[\protect\citeauthoryear{Ekman}{2003}]{ekman2003darwin}
Paul Ekman.
\newblock Darwin, deception, and facial expression.
\newblock {\em Annals of the New York Academy of Sciences}, 1000(1):205--221,
  2003.

\bibitem[\protect\citeauthoryear{K{\"a}chele \bgroup \em et al.\egroup
  }{2015}]{kachele2015multimodal}
Markus K{\"a}chele, Patrick Thiam, Mohammadreza Amirian, Philipp Werner,
  Steffen Walter, Friedhelm Schwenker, and G{\"u}nther Palm.
\newblock Multimodal data fusion for person-independent, continuous estimation
  of pain intensity.
\newblock In {\em Proc. of International Conference on Engineering Applications
  of Neural Networks}, pages 275--285, 2015.

\bibitem[\protect\citeauthoryear{Liu \bgroup \em et al.\egroup
  }{2018}]{liu2018remote}
Si-Qi Liu, Xiangyuan Lan, and Pong~C Yuen.
\newblock Remote photoplethysmography correspondence feature for 3d mask face
  presentation attack detection.
\newblock In {\em Proc. of ECCV}, pages 558--573, 2018.

\bibitem[\protect\citeauthoryear{Lopez-Martinez and
  Picard}{2017}]{lopez2017multi}
Daniel Lopez-Martinez and Rosalind Picard.
\newblock Multi-task neural networks for personalized pain recognition from
  physiological signals.
\newblock In {\em Proc. of the International Conference on Affective Computing
  and Intelligent Interaction Workshops and Demos}, pages 181--184, 2017.

\bibitem[\protect\citeauthoryear{Lopez-Martinez and
  Picard}{2018}]{lopez2018continuous}
Daniel Lopez-Martinez and Rosalind Picard.
\newblock Continuous pain intensity estimation from autonomic signals with
  recurrent neural networks.
\newblock In {\em Proc. of the Annual International Conference of Engineering
  in Medicine and Biology Society}, pages 5624--5627, 2018.

\bibitem[\protect\citeauthoryear{Lucey \bgroup \em et al.\egroup
  }{2011}]{LuceyUNBCdatabase}
Patrick Lucey, Jeffrey~F. Cohn, Kenneth~M. Prkachin, Patricia~E. Solomon, and
  Iain Matthews.
\newblock Painful data: The unbc-mcmaster shoulder pain expression archive
  database.
\newblock In {\em Proc. of FG}, 2011.

\bibitem[\protect\citeauthoryear{Mark \bgroup \em et al.\egroup
  }{2016}]{van2016robust}
van~Gastel Mark, Stuijk Sander, and de~Haan Gerard.
\newblock Robust respiration detection from remote photoplethysmography.
\newblock {\em Biomedical optics express}, 7(12):4941--4957, 2016.

\bibitem[\protect\citeauthoryear{Othman \bgroup \em et al.\egroup
  }{2019}]{othman2019crossdb}
Ehsan Othman, Philipp Werner, Frerk Saxen, Ayoub Al-Hamadi, and Steffen Walter.
\newblock Cross-database evaluation of pain recognition from facial video.
\newblock In {\em Proc. of the International Symposium on Image and Signal
  Processing and Analysis}, pages 181--186, 2019.

\bibitem[\protect\citeauthoryear{Rodriguez \bgroup \em et al.\egroup
  }{2017}]{pau2017deep}
Pau Rodriguez, Guillem Cucurull, Jordi Gonz{\`a}lez, Josep~M. Gonfaus, Kamal
  Nasrollahi, Thomas~B. Moeslund, and F.~Xavier Roca.
\newblock Deep pain: Exploiting long short-term memory networks for facial
  expression classification.
\newblock {\em IEEE Trans. on Cybernetics}, 2017.

\bibitem[\protect\citeauthoryear{Sun and Thakor}{2015}]{sun2015PPG}
Yu~Sun and Nitish Thakor.
\newblock Photoplethysmography revisited: from contact to noncontact, from
  point to imaging.
\newblock {\em IEEE Trans. on Biomedical Engineering}, 63(3):463--477, 2015.

\bibitem[\protect\citeauthoryear{Tavakolian \bgroup \em et al.\egroup
  }{2020}]{tavakolian2020self}
Mohammad Tavakolian, Miguel~B. Lopez, and Li~Liu.
\newblock Self-supervised pain intensity estimation from facial videos via
  statistical spatiotemporal distillation.
\newblock {\em Pattern Recognition Letters}, 140:26--33, 2020.

\bibitem[\protect\citeauthoryear{Terkelsen \bgroup \em et al.\egroup
  }{2005}]{terkelsen2005acute}
Astrid~Juhl Terkelsen, Henning M{\o}lgaard, John Hansen, Ole~Kaeseler Andersen,
  and Troels~Staehelin Jensen.
\newblock Acute pain increases heart rate: differential mechanisms during rest
  and mental stress.
\newblock {\em Autonomic Neuroscience}, 121(1-2):101--109, 2005.

\bibitem[\protect\citeauthoryear{Thiam \bgroup \em et al.\egroup
  }{2019}]{thiam2019exploring}
Patrick Thiam, Peter Bellmann, Hans~A Kestler, and Friedhelm Schwenker.
\newblock Exploring deep physiological models for nociceptive pain recognition.
\newblock {\em Sensors}, 19(20):4503, 2019.

\bibitem[\protect\citeauthoryear{Walter \bgroup \em et al.\egroup
  }{2013}]{walter2013biovid}
Steffen Walter, Sascha Gruss, Hagen Ehleiter, Junwen Tan, Harald~C Traue,
  Philipp Werner, Ayoub Al-Hamadi, Stephen Crawcour, Adriano~O Andrade, and
  Gustavo~Moreira da~Silva.
\newblock The biovid heat pain database data for the advancement and systematic
  validation of an automated pain recognition system.
\newblock In {\em Proc. of the International Conference on Cybernetics}, pages
  128--131, 2013.

\bibitem[\protect\citeauthoryear{Werner \bgroup \em et al.\egroup
  }{2014}]{werner2014automatic}
Philipp Werner, Ayoub Al-Hamadi, Robert Niese, Steffen Walter, Sascha Gruss,
  and Harald~C Traue.
\newblock Automatic pain recognition from video and biomedical signals.
\newblock In {\em Proc. of ICPR}, pages 4582--4587, 2014.

\bibitem[\protect\citeauthoryear{Werner \bgroup \em et al.\egroup
  }{2016}]{werner2016automatic}
Philipp Werner, Ayoub Al-Hamadi, Kerstin Limbrecht-Ecklundt, Steffen Walter,
  Sascha Gruss, and Harald~C Traue.
\newblock Automatic pain assessment with facial activity descriptors.
\newblock {\em IEEE Trans. on Affective Computing}, 8(3):286--299, 2016.

\bibitem[\protect\citeauthoryear{Yang \bgroup \em et al.\egroup
  }{2016}]{yang2016pain}
Ruijing Yang, Shujun Tong, Miguel Bordallo, Elhocine Boutellaa, Jinye Peng,
  Xiaoyi Feng, and Abdenour Hadid.
\newblock On pain assessment from facial videos using spatio-temporal local
  descriptors.
\newblock In {\em Proc. of IPTA}, pages 1--6, 2016.

\bibitem[\protect\citeauthoryear{Yi \bgroup \em et al.\egroup
  }{2014}]{yi2014casia}
Dong Yi, Zhen Lei, Shengcai Liao, and Stan~Z Li.
\newblock Learning face representation from scratch.
\newblock {\em arXiv preprint arXiv:1411.7923}, 2014.

\bibitem[\protect\citeauthoryear{Zhang \bgroup \em et al.\egroup
  }{2016}]{zhang2016mtcnn}
Kaipeng Zhang, Zhanpeng Zhang, Zhifeng Li, and Yu~Qiao.
\newblock Joint face detection and alignment using multitask cascaded
  convolutional networks.
\newblock {\em Signal Processing Letters}, 23(10):1499--1503, 2016.

\bibitem[\protect\citeauthoryear{Zhi and Yu}{2019}]{zhi2019multi}
Ruicong Zhi and Junwei Yu.
\newblock Multi-modal fusion based automatic pain assessment.
\newblock In {\em Proc. of the Joint International Information Technology and
  Artificial Intelligence Conference}, pages 1378--1382, 2019.

\end{thebibliography}

\end{document}